\documentclass[conference]{IEEEtran}

\usepackage[bookmarks]{hyperref}
\usepackage{times}
\usepackage[numbers]{natbib}
\usepackage{multicol}
\usepackage{multirow}
\usepackage{booktabs,dcolumn}
\usepackage{amsmath,amsfonts,amssymb}
\usepackage{algorithm, algorithmic}
\usepackage{graphicx}
\usepackage{caption}
\usepackage{subcaption}
\usepackage{url}
\usepackage{cleveref}
\usepackage{bm}
\usepackage{color}

\crefformat{footnote}{#2\footnotemark[#1]#3}
\makeatletter
\renewcommand*\env@matrix[1][*\c@MaxMatrixCols c]{
  \hskip -\arraycolsep
  \let\@ifnextchar\new@ifnextchar
  \array{#1}}
\makeatother
\DeclareMathOperator*{\argmin}{arg\!\min}
\DeclareMathOperator*{\argmax}{arg\!\max}

\begin{document}

\title{Incremental Sparse GP Regression for Continuous-time Trajectory Estimation \& Mapping }

\author{
\IEEEauthorblockN{Xinyan Yan} 
\IEEEauthorblockA{College of Computing\\
Georgia Institute of Technology\\
Atlanta, GA 30332, USA \\
\texttt{xinyan.yan@cc.gatech.edu}}
\and
\IEEEauthorblockN{Vadim Indelman}
\IEEEauthorblockA{Department of Aerospace Engineering\\
Technion - Israel Institute of Technology\\
 Haifa, 32000, Israel \\
\texttt{vadim.indelman@technion.ac.il}}
\and
\IEEEauthorblockN{Byron Boots}
\IEEEauthorblockA{College of Computing\\
Georgia Institute of Technology\\
Atlanta, GA 30332, USA \\
\texttt{bboots@cc.gatech.edu} }
}

\maketitle

\begin{abstract}
Recent work on simultaneous trajectory estimation and mapping (STEAM) for mobile robots has found success by representing the trajectory as a Gaussian process. Gaussian processes can represent a continuous-time trajectory, elegantly handle asynchronous and sparse measurements, and allow the robot to query the trajectory to recover its estimated position at any time of interest. A major drawback of this approach is that STEAM is formulated as a \emph{batch} estimation problem.  In this paper we provide the critical extensions necessary to transform the existing batch algorithm into an extremely efficient incremental algorithm. In particular, we are able to vastly speed up the solution time through efficient variable reordering and incremental sparse updates, which we believe will greatly increase the practicality of Gaussian process methods for robot mapping and localization. Finally, we demonstrate the approach and its advantages on both synthetic and real datasets.
\end{abstract}

\IEEEpeerreviewmaketitle

\section{Introduction \& Related Work}
Simultaneously recovering the location of a robot and a map of its environment  from sensor readings is a fundamental challenge in robotics~\cite{Thrun2005,Durrant-Whyte06,Bailey06}.

Well-known approaches to this problem, such as square root smoothing and mapping (SAM)~\cite{Dellaert-SRSAM}, have focused on  regression-based methods that exploit the sparse structure of the problem to efficiently compute a solution. The main weakness of the original SAM algorithm was that it was a \emph{batch} method: all of the data must be collected before a solution can be found. For a robot traversing an environment, the inability to update an estimate of its trajectory online is a significant drawback. In response to this weakness, \citet{Kaess-iSAM} developed a critical extension to the batch SAM algorithm, incremental smoothing and mapping (iSAM),  that overcomes this problem by \emph{incrementally} computing a solution. The main drawback of iSAM, was that the approach required costly periodic batch steps for variable reordering to maintain sparsity and relinearization. This approach was extended in iSAM~2.0~\cite{Kaess-iSAM2}, which employs an efficient data structure called the \emph{Bayes tree}~\cite{kaess2011bayes} to perform incremental variable reordering and just-in-time relinearization, thereby eliminating the bottleneck caused by batch variable reordering and relinearization. The iSAM~2.0 algorithm and its extensions are widely considered to be state-of-the-art in robot trajectory estimation and mapping.

The majority of previous approaches to trajectory estimation and mapping, including the smoothing-based SAM family of algorithms, have formulated the problem in discrete time~\cite{Montemerlo02, Thrun2005,Durrant-Whyte06,Bailey06,Dellaert-SRSAM,Kaess-iSAM2,Boots-spectralROSLAM}. However, discrete-time representations are restrictive: they are not easily extended to trajectories with irregularly spaced waypoints or asynchronously sampled measurements. A continuous-time formulation of the SAM problem where measurements constrain the trajectory at any point in time, would elegantly contend with these difficulties.  Viewed from this perspective, the robot trajectory is a \emph{function} $\bm{x}(t)$, that maps any time $t$ to a robot state. The problem of estimating this function along with landmark locations has been dubbed \emph{simultaneous trajectory estimation and mapping} (STEAM) \cite{Barfoot-RSS-14}. 

Tong et al.~\cite{tong2013gaussian} proposed a Gaussian process (GP) regression approach to solving the STEAM problem. While their approach was able to accurately model and interpolate asynchronous data to recover a trajectory and landmark estimate, it suffered from significant computational challenges: naive Gaussian process approaches to regression have notoriously high space and time complexity. Additionally, Tong et al.'s approach is a \emph{batch} method, so updating the solution necessitates saving all of the data and completely resolving the problem. In order to combat the computational burden, Tong et al.'s approach was extended in Barfoot et al.~\cite{Barfoot-RSS-14} to take advantage of the sparse structure inherent in the STEAM problem. The resulting algorithm significantly speeds up solution time and can be viewed as a continuous-time analog of Dellaert's original square-root SAM algorithm~\cite{Dellaert-SRSAM}. Unfortunately, like SAM, Barfoot et al.'s GP-based algorithm remains a batch algorithm, which is a disadvantage for robots that need to continually update the estimate of their trajectory and environment.

In this work, we provide the critical extensions necessary to transform the existing Gaussian process-based approach to solving the STEAM problem into an extremely efficient incremental approach. Our algorithm elegantly combines the benefits of Gaussian processes and iSAM~2.0. Like the GP regression approaches to STEAM, our approach can model continuous trajectories, handle asynchronous measurements, and naturally interpolate states to speed up computation and reduce storage requirements,  and, like iSAM 2.0, our approach uses a Bayes tree to efficiently calculate a \emph{maximum a posteriori} (MAP) estimate of the GP trajectory while performing incremental factorization, variable reordering, and just-in-time relinearization. The result is an online GP-based solution to the STEAM problem that remains computationally efficient while scaling up to large datasets.

\section{Batch Trajectory Estimation \& Mapping as Gaussian Process Regression}\label{sec_trajest}
We begin by describing how the simultaneous trajectory estimation and mapping (STEAM) problem can be formulated in terms of Gaussian process regression. Following Tong et al.~\cite{tong2013gaussian} and Barfoot et al.~\cite{Barfoot-RSS-14}, we represent robot trajectories 
as functions of time $t$ sampled from a Gaussian process:
\begin{align} \label{eqn_process_model}
  \bm{x}(t) &\sim \mathcal{GP}(\bm{\mu}(t), \bm{\mathcal{K}}(t, t')), \hspace{12pt} t_0 < t,t'
\end{align}

Here, $\bm{x}(t)$ is the continuous-time trajectory of the robot through state-space, represented by a Gaussian process with mean $\bm{\mu}(t)$ and covariance $\bm{\mathcal{K}}(t, t')$ functions. 

We next define a finite set of measurements: 
\begin{align} \label{eqn_measurement_model}
  \bm{y}_i &= \bm{h}_i(\bm{\theta}_i) + \bm{n}_i,\hspace{8pt} \bm{n}_i \sim \mathcal{N}(\bm{0}, \bm{R}_i), \hspace{8pt} i = 1, 2, ..., N
\end{align}

The measurement $\bm{y}_i$ can be any linear or nonlinear functions of a set of related variables $\bm{\theta}_i$ plus some Gaussian noise $\bm{n}_i$. The related variables for a range measurement are the robot state at the corresponding measurement time $\bm{x}(t_i)$ and the associated landmark location $\bm{\ell}_j$. We assume the total number of measurements is $N$, and the number of trajectory states at measurement times are $M$.

Based on the definition of Gaussian processes, any finite collection of robot states has a joint Gaussian distribution~\cite{Rasmussen06Book}. So the robot states at measurement times are normally distributed with mean $\bm{\mu}$ and covariance $\bm{\mathcal{K}}$.
\begin{equation}
\begin{split}
&\bm{x} \sim \mathcal{N}(\bm{\mu}, \bm{\mathcal{K}}), \hspace{12pt} \bm{x} = [\begin{array}{ccc}  \bm{x}(t_1)^{\intercal}  &   \hdots & \bm{x}(t_M)^{\intercal}\end{array}]^\intercal\\
&\bm{\mu} =  [\begin{array}{ccc}  \bm{\mu}(t_1)^{\intercal}  &   \hdots & \bm{\mu}(t_M)^{\intercal}\end{array}]^\intercal, \hspace{12pt}\bm{\mathcal{K}}_{ij} = \bm{\mathcal{K}}(t_i, t_j)
\end{split}
\end{equation}

Note that any point along the continuous-time trajectory can be estimated from the Gaussian process model. Therefore, the trajectory does not need to be discretized and robot trajectory states do not need to be evenly spaced in time, which is an advantage of the Gaussian process approach over discrete-time approaches (e.g.~Dellaert's square-root SAM~\cite{Dellaert-SRSAM}).

The landmarks $\bm{\ell}$ which represent the map are assumed to conform to a joint Gaussian distribution with mean $\bm{d}$ and covariance $\bm{W}$ (Eq.~\ref{eqn_landmark_dist}). The prior distribution of the combined state $\bm{\theta}$ that consists of robot trajectory states at measurement times and landmarks is, therefore, a joint Gaussian distribution (Eq.~\ref{eqn_prior_dist}).
\begin{align} 
&\bm{\ell} \sim \mathcal{N}(\bm{d}, \bm{W}), \hspace{12pt} \bm{\ell} = [\begin{array}{cccc} \bm{\ell}_1^\intercal & \bm{\ell}_2^\intercal & \hdots& \bm{\ell}_O^\intercal \end{array}]^\intercal \label{eqn_landmark_dist}\\
&\bm{\theta} \sim \mathcal{N}(\bm{\eta}, \bm{\mathcal{P}}), \hspace{12pt}\bm{\eta} = [\begin{array}{cc}  \bm{\mu}^\intercal &  \bm{d}^\intercal\end{array}]^\intercal, \hspace{12pt}
\bm{\mathcal{P}} = 
\begin{bmatrix}
  \bm{\mathcal{K}} & \\
  & \bm{W} 
\end{bmatrix}    \label{eqn_prior_dist}
\end{align}
To solve the STEAM problem, given the prior distribution of the combined state and the likelihood of measurements, we compute the \emph{maximum a posteriori} (MAP) estimate of the combined state \emph{conditioned} on measurements via Bayes' rule: 
\begin{align}\label{eqn_map}
  \bm{\theta}^* &\triangleq \hat{\bm{\theta}}_{MAP} =  \argmax\limits_{\bm{\theta}} \, p(\bm{\theta} | \bm{y}) = \argmax\limits_{\bm{\theta}} \, \frac{ p(\bm{\theta}) p( \bm{y}|\bm{\theta})}{p(\bm{y})}\nonumber  \\
  &= \argmax\limits_{\bm{\theta}} p(\bm{\theta})p(\bm{y} |\bm{\theta}) = \argmin\limits_{\bm{\theta}}  \left(-\log p(\bm{\theta}) - \log p(\bm{y} |\bm{\theta}) \right)\nonumber \\
  &= \argmin\limits_{\bm{\theta}}\, \left( \| \bm{\theta} - \bm{\eta} \|_{\bm{\mathcal{P}}}^{2} + \| \bm{h}(\bm{\theta}) - \bm{y} \|_{\bm{R}}^{2} \right)
\end{align}
where the norms are Mahalanobis norms  defined as: $\|\bm{e}\|^2_{\bm{\Sigma}} = \bm{e}^{\intercal} \bm{\Sigma}^{-1} \bm{e}$, and $\bm{h}(\bm{\theta})$ and $\bm{R}$ are the mean and covariance of the measurements collected, respectively:
\begin{align}
\bm{h}(\bm{\theta}) &= [\begin{array}{cccc}
  \bm{h}_1(\bm{\theta}_1) &
  \bm{h}_2(\bm{\theta}_2) &
  \hdots &
  \bm{h}_N(\bm{\theta}_N)
\end{array}]^\intercal\\
\bm{R} &= \text{diag}(\bm{R}_1, \bm{R}_2, \hdots, \bm{R}_N)
\end{align}
Because both covariance matrices $\bm{\mathcal{P}}$ and $\bm{R}$ are positive definite, the objective in Eq.~\ref{eqn_map} corresponds to a least squares problem. Consequently, if some of the measurement functions $\bm{h}_i(\cdot)$ are nonlinear, this becomes a nonlinear least squares problem, in which case iterative methods including Gauss-Newton and Levenberg-Marquardt~\cite{Dennis-1996} can be utilized. A linearization of a measurement function at current state estimate $\bar{\bm{\theta}}_i$ can be accomplished by a first-order Taylor expansion:
\begin{equation} \label{eqn_measurement_linearization}
  \bm{h}_i \left( \bar{\bm{\theta}}_i + \delta\bm{\theta}_i \right) \approx \bm{h}_i(\bar{\bm{\theta}}_i) +  
  \frac {\partial \bm{h}_i} { \partial {\bm{\theta}_i}}\bigg|_{\bar{\bm{\theta}}_i} \thinspace \delta\bm{\theta}_i
\end{equation}
Combining Eq.~\ref{eqn_measurement_linearization} with Eq.~\ref{eqn_map}, the optimal increment $\delta\bm{\theta}^*$ at the current combined state estimate $\bar{\bm{\theta}}$ is 
\begin{align}\label{eqn_map_linearized}
\delta\bm{\theta}^* \hspace{-.5mm} =\hspace{-.5mm} \argmin\limits_{\delta\bm{\theta}}\, \left( \| \bar{\bm{\theta}} \hspace{-.5mm}+\hspace{-.5mm} \delta{\bm{\theta}} \hspace{-.5mm}- \hspace{-.5mm}\bm{\eta} \|_{\bm{\mathcal{P}}}^{2} + \| \bm{h}(\bar{\bm{\theta}})\hspace{-.5mm} +\hspace{-.5mm} \bm{H}\delta{\bm{\theta}} \hspace{-.5mm}-\hspace{-.5mm} \bm{y} \|_{\bm{R}}^{2} \right)
\end{align}
Where $\bm{H}$ is the measurement Jacobian matrix:
\begin{align}
\bm{H} = \text{diag}(\bm{H}_1, \bm{H}_2, \hdots, \bm{H}_N)
, \hspace{20pt}
\bm{H}_i = \frac { \partial \bm{h}_i } {\partial {\bm{\theta}_i}} \bigg |_{  {\bar{\bm{\theta}}}_i}
\end{align}
To solve the linear least squares problem in Eq.~\ref{eqn_map_linearized}, we take the derivative with respect to $\delta\bm{\theta}$, and set it to zero, which gives us $\delta\bm{\theta}^*$ embedded in a set of linear equations
\begin{align} \label{eqn_linear_equations}
\hspace{-1mm}   \underbrace{(\bm{\mathcal{P}}^{-1}\hspace{-1.5mm} +\hspace{-.5mm} \bm{H}^\intercal \bm{R}^{-1}\hspace{-.5mm}\bm{H})}_{\bm{\mathcal{I}}} \delta\bm{\theta}^* \hspace{-.5mm}= \hspace{-.5mm}\underbrace{ \bm{\mathcal{P}}^{-1}\hspace{-.5mm}(\bm{\eta} \hspace{-.5mm}- \hspace{-.5mm}\bar{\bm{\theta}}) \hspace{-.5mm}+\hspace{-.5mm} \bm{H}^\intercal \bm{R}^{-1} (\bm{y}\hspace{-.5mm}-\hspace{-.5mm}\bar{\bm{h}})}_{\bm{b}}\hspace{-1mm}
\end{align}
with covariance
\begin{align}
\mathrm{cov}(\delta{\bm{\theta}}^*, \delta{\bm{\theta}}^*) = \bm{\mathcal{I}}^{-1}
\end{align}
The positive definite matrix $\bm{\mathcal{P}}^{-1} + \bm{H}^\intercal \bm{R}^{-1}\bm{H}$ is the \emph{a posteriori} information matrix, which we label $\bm{\mathcal{I}}$.
To solve this set of linear equations for $\delta\bm{\theta}^*$, we do not actually have to calculate the inverse $\bm{\mathcal{I}}^{-1}$. Instead, factorization-based methods can provide a fast, numerically stable solution. For example, $\delta\bm{\theta}^*$ can be found by first performing a Cholesky factorization $\bm{\mathcal{L}}\bm{\mathcal{L}}^\intercal = \bm{\mathcal{I}}$, and then solving $\bm{\mathcal{L}}\bm{d} = \bm{b}$ and $\bm{\mathcal{L}}^\intercal \delta\bm{\theta}^* = \bm{d}$ by back substitution. At each iteration we perform a \emph{batch} state estimation update $\bar {\bm{\theta}} \leftarrow \bar {\bm{\theta}} + \delta\bm{\theta}^*$ and repeat the process until convergence.

If $\bm{\mathcal{I}}$ is dense, the time complexity of a Cholesky factorization and back substitution are $O(n^3)$ and $O(n^2)$ respectively, where $\bm{\mathcal{I}} \in \mathbb{R}^{n \times n}$~\cite{Golub:1996}. However, if $\bm{\mathcal{I}}$ has sparse structure, then the solution can be found much faster. For example, for a narrowly banded matrix, the computation time is $O(n)$ instead of $O(n^3)$~\cite{Golub:1996}. Fortunately, we can guarantee sparsity for the STEAM problem (see Section~\ref{subsec_sparse_GP_regression} below).

\subsection{State Interpolation}\label{subsec_state_interpolation}

An advantage of the Gaussian process representation of the robot trajectory is that any trajectory state can be interpolated from other states by computing the posterior mean~\cite{tong2013gaussian}:
\begin{align} \label{eqn_batch_query}
\bar{\bm{x}}(t) = \bm{\mu}(t) + \bm{\mathcal{K}}(t) \bm{\mathcal{K}}^{-1}(\bar{\bm{x}} - \bm\mu), 
\end{align}
with 
\begin{align}
\bar{\bm{x}} &= [\begin{array} {ccc} \bar{\bm{x}}(t_1) & \hdots &  \bar{\bm{x}}(t_M)  \end{array}]^\intercal \quad \text{and}\nonumber\\
\bm{\mathcal{K}}(t) &= [\begin{array}{ccc} \bm{\mathcal{K}}(t, t_1) & \hdots & \bm{\mathcal{K}}(t, t_M) \end{array}].
\end{align}
By utilizing interpolation, we can reduce the number of robot trajectory states that we need to estimate in the optimization procedure~\cite{tong2013gaussian}.
For simplicity, assume $\bm{\theta}_i$, the set of the related variables of the $i$th measurement according to the model (Eq.~\ref{eqn_measurement_model}), is $\bm{x}(t_j)$. Then, after interpolation, Eq.~\ref{eqn_measurement_linearization} becomes:
\begin{align} \label{eqn_interpolated_measurement_linearization}
&\bm{h}_i \left( \bar{\bm{\theta}}_i + \delta\bm{\theta}_i \right) = \bm{h}_i \left( \bar{\bm{x}}(t_j) + \delta\bm{x}(t_j) \right)\nonumber \\
&\approx \bm{h}_i(\bar{\bm{x}}(t_j))+  
\frac {\partial \bm{h}_i} { \partial {\bm{x}(t_j)}}  \cdot 
\frac {\partial \bm{x}(t_j)} { \partial {\bm{x}}} \thinspace \bigg|_{\bar{\bm{x}}}
\delta{\bm{x}}\nonumber \\
&= \bm{h}_i\hspace{-.5mm}\left(\bm{\mu}(t_j) \hspace{-.5mm}+ \hspace{-.5mm}\bm{\mathcal{K}}(t_j) \bm{\mathcal{K}}^{-1}(\bar{\bm{x}} \hspace{-.5mm}-\hspace{-.5mm} \bm\mu)\right) \hspace{-.5mm}+\hspace{-.5mm} \bm{H}_i \bm{\mathcal{K}}(t_j) \bm{\mathcal{K}}^{-1} \delta{\bm{x}}
\end{align}
By employing Eq.~\ref{eqn_interpolated_measurement_linearization} during optimization, we can make use of measurement $i$ without explicitly estimating the trajectory states that it relates to. We exploit this advantage to greatly speed up the solution to the STEAM problem in practice (Section~\ref{sec_experiment}).

\subsection{Sparse Gaussian Process Regression}\label{subsec_sparse_GP_regression}

The efficiency of the Gaussian Process Gauss-Newton algorithm presented in Section~\ref{sec_trajest} is heavily dependent on the choice of kernel. It is well-known that if the information matrix $\bm{\mathcal{I}}$ is sparse, then it is possible to very efficiently compute the solution to Eq.~\ref{eqn_linear_equations}~\cite{Dellaert-SRSAM}. 
Barfoot et al. suggest a kernel matrix with a sparse inverse that is well-suited to the simultaneous trajectory estimation and mapping problem~\cite{Barfoot-RSS-14}. 
In particular, Barfoot et al.  show that $\bm{\mathcal{K}}^{-1}$ is exactly block-tridiagonal when the GP is assumed to be generated by linear, time-varying (LTV) stochastic differential equation (SDE) which we describe here: 
\begin{align} \label{eqn_sde}
\dot{\bm{x}}(t) &= \bm{A}(t)\bm{x}(t) + \bm{v}(t) + \bm{F}(t)\bm{w}(t),\\
\bm{w}(t) &\sim  \mathcal{GP}(\bm{0}, \thinspace \bm{Q}_c\delta(t-t')) \hspace{12pt} t_0 < t,t'
\end{align}
where $\bm{x}(t)$ is trajectory, $\bm{v}(t) $ is known exogenous input, $\bm{w}(t)$ is process noise, and $\bm{F}(t)$ is time-varying system matrix. The process noise $\bm{w}(t)$ is modeled by a Gaussian process, and $\delta(\cdot)$ is the \emph{Dirac delta function}. (See~\cite{Barfoot-RSS-14} for details). We consider a specific case of this model in the experimental results in Section~\ref{subsec_synthetic_experiment}.

\begin{figure}
 \centering
 \begin{subfigure}{.24\textwidth}
   \centering
   \includegraphics[width=\linewidth]{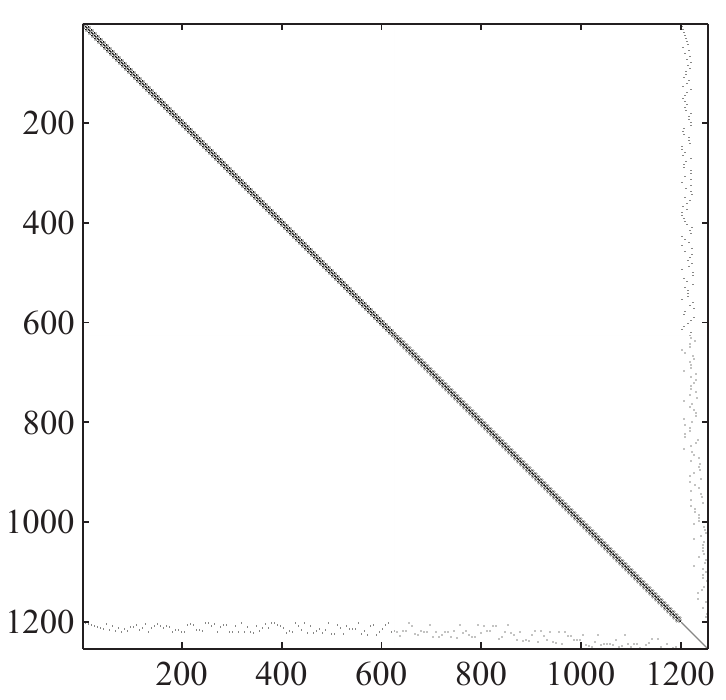}
   \caption{XL ordering\cref{foot_XL_ordering}}
   \label{fig_original_info_mat}
 \end{subfigure}
 \begin{subfigure}{.24\textwidth}
   \centering
   \includegraphics[width=\linewidth]{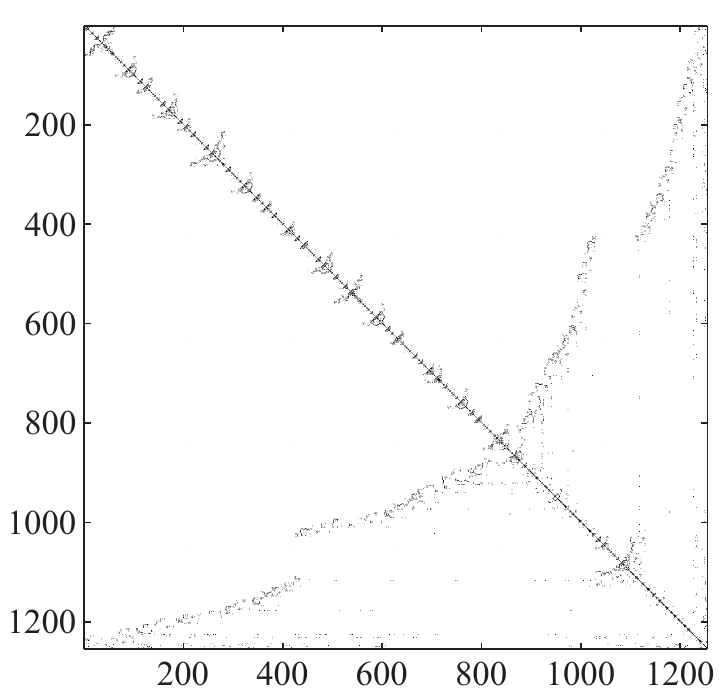}
   \caption{SYMAMD ordering\cref{foot_SYMAMD_ordering}}
   \label{fig_reordered_info_mat}
 \end{subfigure}
 \caption{Sparse information matrices. The information matrix $\bm{\mathcal{I}}$ with XL ordering\cref{foot_XL_ordering} and SYMAMD ordering\cref{foot_SYMAMD_ordering}. Both sparse matrices have the same number of non-zero elements, yet the second matrix can be factored much more efficiently due to the heuristic ordering of the matrix columns. (See Table~\ref{tab_chol}). For  illustration, only 200 trajectory states are shown here.}
 \label{fig_info_mat}
 \end{figure}

Assuming the GP is generated by Eq.~\ref{eqn_sde}, the measurements are landmark and odometry measurements, and the variables are ordered in XL ordering\footnote{\label{foot_XL_ordering} XL ordering is an ordering where process variables come before landmarks variables.}, the sparse information matrix becomes
\begin{align} \label{eqn_sparseI}
\bm{\mathcal{I}} = \left [ \begin{array}{cc}\bm{\mathcal{I}}_{xx}& \bm{\mathcal{I}}_{x\ell} \\ \bm{\mathcal{I}}_{x\ell}^\intercal & \bm{\mathcal{I}}_{\ell\ell} \end{array}\right ]\end{align} 
where $\bm{\mathcal{I}}_{xx}$ is block-tridiagonal and $\bm{\mathcal{I}}_{\ell\ell}$ is block-diagonal. $\bm{\mathcal{I}}_{x\ell}$'s  density depends on the frequency of landmark measurements, and how they are taken. See Fig.~\ref{fig_original_info_mat} for an example.

When the GP is generated by LTV SDE, Barfoot et al. prove that $\bm{\mathcal{K}}(t) \bm{\mathcal{K}}^{-1}$ in Eq.~\ref{eqn_batch_query} has a specific sparsity pattern, only two column blocks that correspond to trajectory states at $t_i$ and $t_{i+1}$, where $t_i < t < t_{i+1}$, are nonzero. In other words, $\bar{\bm{x}}(t)$ is an affine function of only two nearby states $\bar{\bm{x}}(t_i)$ and $\bar{\bm{x}}(t_{i+1})$:
\begin{align} \label{eqn_sparse_query}
\bar{\bm{x}}(t) =& \bm{\mu}(t) + \bm{\Lambda}(t) \left( \bar{\bm{x}}(t_i) \hspace{-.5mm} -\hspace{-.5mm}  \bm{\mu}(t_i)\right) \hspace{-.5mm} +\hspace{-.5mm}  \bm{\Psi}(t) \left(\bar{\bm{x}}(t_{i+1}) \hspace{-.5mm} - \hspace{-.5mm} \bm{\mu}(t_{i+1})\right),\nonumber\\
&t_i < t < t_{i+1}
\end{align}
Thus, it only takes $O(1)$ time to query any $\bar{\bm{x}}(t)$ using Eq.~\ref{eqn_sparse_query}. Moreover, because interpolation of a state is only determined by the two nearby states, measurement interpolation in Eq.~\ref{eqn_interpolated_measurement_linearization} can be significantly simplified.

\section{Batch GP-Regression with Variable Reordering} \label{sec_var_reordering}
Previous work on batch continuous-time trajectory estimation as sparse Gaussian process regression~\cite{tong2013gaussian,Barfoot-RSS-14} assumes that the information matrix $\bm{\mathcal{I}}$ is sparse (Eq.~\ref{eqn_sparseI}) and applies standard block elimination to factor and solve Eq.~\ref{eqn_linear_equations}. Despite the sparsity of $\bm{\mathcal{I}}$, for large numbers of landmarks this process can be very inefficient. Inspired by square root SAM~\cite{Dellaert-SRSAM}, which uses variable reordering for efficient Cholesky factorization in a discrete-time context, 
we show that factorization-time can be dramatically improved by matrix column reordering in the sparse Gaussian process context as well.

It is reasonable to base our approach on SAM because 
the information matrix and factor graph of the sparse GP~\cite{Barfoot-RSS-14} has structure similar to the SAM formulations of the problem~\cite{Dellaert-SRSAM, Kaess-iSAM}, and the intuitions from previous discrete-time approaches apply here. If the Cholesky decompositions are performed naively, fill-in can occur, where entries that are zero in the information matrix become non-zero in the Cholesky factor. This occurs because the Cholesky factor of a sparse matrix is guaranteed to be sparse for some variable orderings, but not all variable orderings~\cite{Ranganathan-Online-Sparse}. Therefore, we want to find a good variable ordering so that the Cholesky factor is sparse.

\begin{figure}
\centering
\begin{subfigure}{.24\textwidth}
  \centering
  \includegraphics[width=\linewidth]{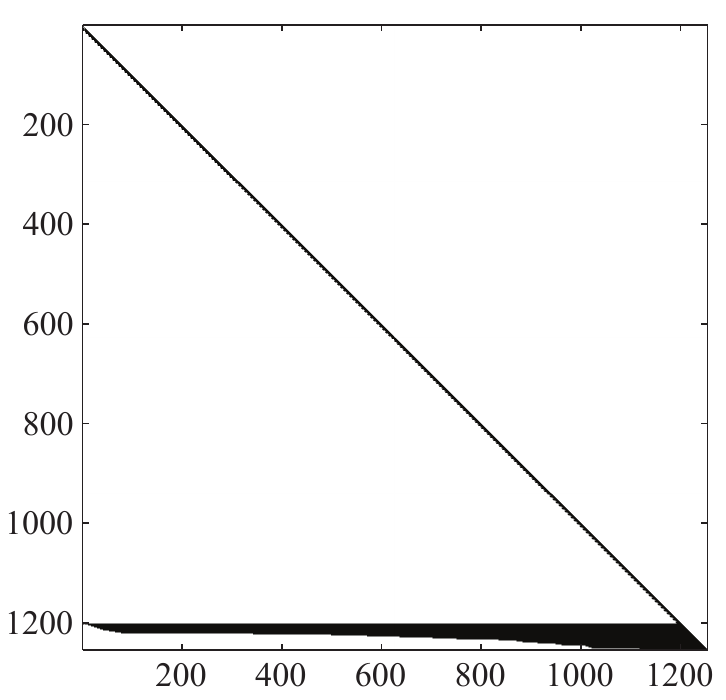}
  \caption{XL ordering\cref{foot_XL_ordering}}
  \label{fig_chol_original}
\end{subfigure}
\begin{subfigure}{.24\textwidth}
  \centering
  \includegraphics[width=\linewidth]{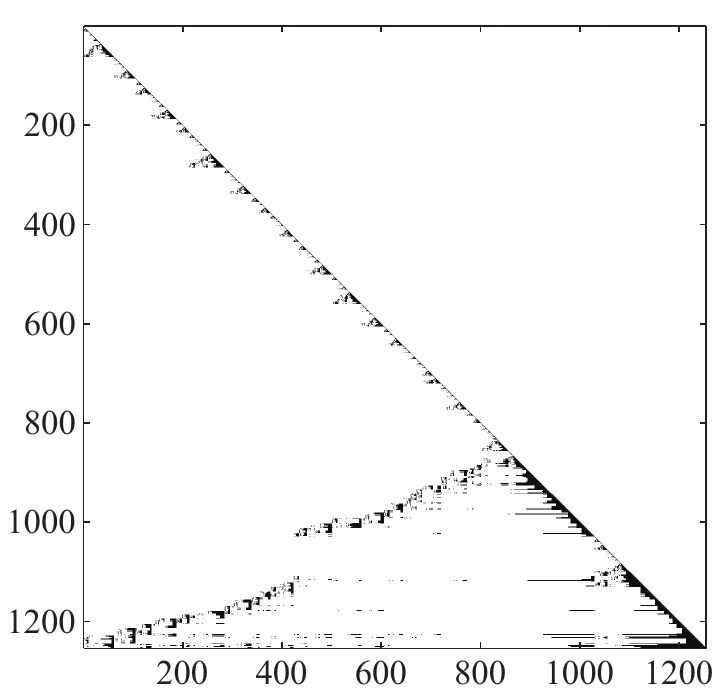}
  \caption{SYMAMD ordering\cref{foot_SYMAMD_ordering}}
  \label{fig_chol_reordered}
\end{subfigure}
\caption{The Cholesky factors $\bm{\mathcal{L}}$ of $\bm{\mathcal{I}}$. In (a), $\bm{\mathcal{L}}$ is computed with XL ordering\cref{foot_XL_ordering}, which exhibits fill-in. When computed with SYMAMD ordering in (b), $\bm{\mathcal{L}}$ is more sparse. For illustration, only 200 states are shown here.} 
\label{fig_chol}
\end{figure}

Although finding the optimal ordering for a symmetric positive definite matrix is NP-complete~\cite{Yannakakis:1981}, good heuristics do exist. One such heuristic is symmetric approximate minimum degree permutation (SYMAMD)\footnote{\label{foot_SYMAMD_ordering}SYMAMD is a variant of column approximate minimum degree ordering (COLAMD)~\cite{Davis:2004} on positive definite matrix.}~\cite{Davis:2004}. To demonstrate the benefits of variable reordering, we constructed a synthetic example and compared different approaches. The example, which is explained in detail in Section~\ref{subsec_synthetic_experiment}, consists of 1,500 time steps with trajectory states, $\bm{x}(t_i) = [\begin{array}{cc} \bm{p}(t_i) & \dot{\bm{p}}(t_i) \end{array}]^\intercal$, $\bm{p}(t_i) = [\begin{array}{ccc} x(t_i) & y(t_i) & \theta(t_i) \end{array}]^\intercal$, and with odometry and range measurements. The total number of landmarks is 298. The structure of the information matrix $\bm{\mathcal{I}}$ and Cholesky factor $\bm{\mathcal{L}}$, with and without variable reordering, are compared in Fig.~\ref{fig_info_mat} and Fig.~\ref{fig_chol}. Although variable reordering does not change the sparsity of the information matrix $\bm{\mathcal{I}}$ (Fig.~\ref{fig_info_mat}), it dramatically increases the sparsity of the Cholesky factor $\bm{\mathcal{L}}$ (Fig.~\ref{fig_chol}). Table~\ref{tab_chol} demonstrates this clear benefit of reordering. The Cholesky factor after SYMAMD ordering contains 10.6\% non-zeroes of XL ordering~\ref{foot_XL_ordering}, and takes 2.83\% of the time, which are significant improvements in both time and space complexity. 

We also experimented with block SYMAMD~\cite{Dellaert-SRSAM}, which exploits domain knowledge to group together variables belonging to a particular trajectory state $\bm{x}(t_i)$ or landmark location $\bm{\ell}_j$  before performing SYMAMD and empirically further improves performance.  

\begin{table}[t]\small
\caption{Cost of Cholesky factorization with different ordering methods including ordering time}
\label{tab_chol}
\begin{center}
\begin{tabular}{l |l |l |l}
& \multicolumn{1}{c|}{\bf XL\cref{foot_XL_ordering}}  &\multicolumn{1}{c|}{\bf SYMAMD} & \multicolumn{1}{c}{\bf Block SYMAMD} \\
\hline \hline
nnz \footnotemark  & 1817499 & 192285 & 176105 \\
\hline
time (sec) & 0.967720 & 0.027402 & 0.017500
\end{tabular}
\end{center}
\end{table}

It is straightforward to incorporate variable reordering methods like SYMAMD and block SYMAMD into the batch GP-Regression algorithm from Section~\ref{sec_trajest}. Given a new batch of data, directly update the sparse information matrix $\bm{\mathcal{I}}$, reorder the variables with (block) SYMAMD, and then recompute the Cholesky factor $\bm{\mathcal{L}}$ on the way to solving for $\delta\bm{\theta}$ in Eq.~\ref{eqn_linear_equations}. 

In most STEAM problems, we are interested in estimating the robot's trajectory \emph{as it traverses the environment}. 
In Alg.~\ref{alg_incremental_gpgn}, we accomplish this by repeatedly executing the batch algorithm with variable reordering. Although this approach seems like it should be very costly, with variable reordering this method it is actually quite efficient. Building and factoring the sparse information matrix is much faster than the linearization step required for a single iteration of the Gauss-Newton algorithm. Since the computational bottleneck is not the Cholesky decomposition, but rather the relinearization of the measurement model, we suggest only periodic Gauss-Newton iterations.

 \begin{algorithm}\footnotesize
 \caption{Periodic Batch Sparse GP Regression} \label{alg_incremental_gpgn}
 \begin{algorithmic}
 \WHILE{collecting data}
 \vspace{2pt}
 \STATE 1. Get measurement results belonging to this period, $\bm{y} \leftarrow [\bm{y}, \bm{y}_{new}]^\intercal$
 \vspace{2pt}
 \STATE 2. Initial guess for the newly encountered states, $\bar{\bm{\theta}} \leftarrow  [\bar{\bm{\theta}}, \bar{\bm{\theta}}_{new}]^\intercal$
 \vspace{2pt}
 \STATE 3. Build the measurement Jacobian $\bm{H}$, and then $\bm{\mathcal{I}}$ and $\bm{b}$ required in Eq.~\ref{eqn_linear_equations}
 \STATE 4. Find an ordering \textit{p} for $\bm{\mathcal{I}}$, and reorder $\bm{\mathcal{I}}_p \xleftarrow{p} \bm{\mathcal{I}}$, $\bm{b}_p \xleftarrow{p} \bm{b}$
 \vspace{2pt}
 \STATE 5. Solve $\bm{\mathcal{I}}_p \delta\bm{\theta}_p^* = \bm{b}_p$ using Cholesky factorization 
 \vspace{2pt}
 \STATE 6. Recover the solution $\delta\bm{\theta}^* \xleftarrow{r} \delta\bm{\theta}_p^*$ by inverse ordering $r = p^{-1}$ 
 \vspace{2pt}
 \STATE 7. Update estimate $\bar{\bm{\theta}} \leftarrow \bar{\bm{\theta}} + \delta\bm{\theta}^*$
 \vspace{2pt}
 \ENDWHILE
 \end{algorithmic}
 \end{algorithm}

\footnotetext{\label{foot_nnz}The number of non-zero elements.}

\section{Bayes Tree for Fast Incremental Updates to Sparse GP Regression} \label{sec_incremental_updates}
Despite the efficiency of periodic batch updates, Alg.~\ref{alg_incremental_gpgn} is still repeatedly executing a batch algorithm that requires reordering and refactoring $\bm{\mathcal{I}}$, and periodically relinearizing the measurement function for all of the estimated states each time new data is collected. Here we provide the extensions necessary to avoid these costly steps and turn the naive batch algorithm into an efficient, truly incremental, algorithm. 
The key idea is to perform just-in-time relinearization and to efficiently \emph{update} an existing sparse factorization instead of re-calculating one from scratch.

\subsection{The Bayes Tree Data Structure}
We base our approach on iSAM 2.0 proposed by Kaess et al.~\cite{Kaess-iSAM2}, which was designed to efficiently solve a nonlinear estimation problem in an incremental and real-time manner by directly operating on the factor graph representation of the SAM problem. The core technology behind iSAM 2.0 is the \emph{Bayes tree} data structure which allows for incremental variable reordering and fluid relinearization~\cite{kaess2011bayes}. Demonstrated by Kaess et al.~\cite{Kaess-iSAM2}, Bayes tree provides dramatic speedup compared to batch method, with negligible loss in accuracy. We apply the same data structure to sparse Gaussian process regression in the context of the STEAM problem, thereby eliminating the need for periodic batch computation. 

\begin{figure}
\centering
\includegraphics[width=.9\linewidth]{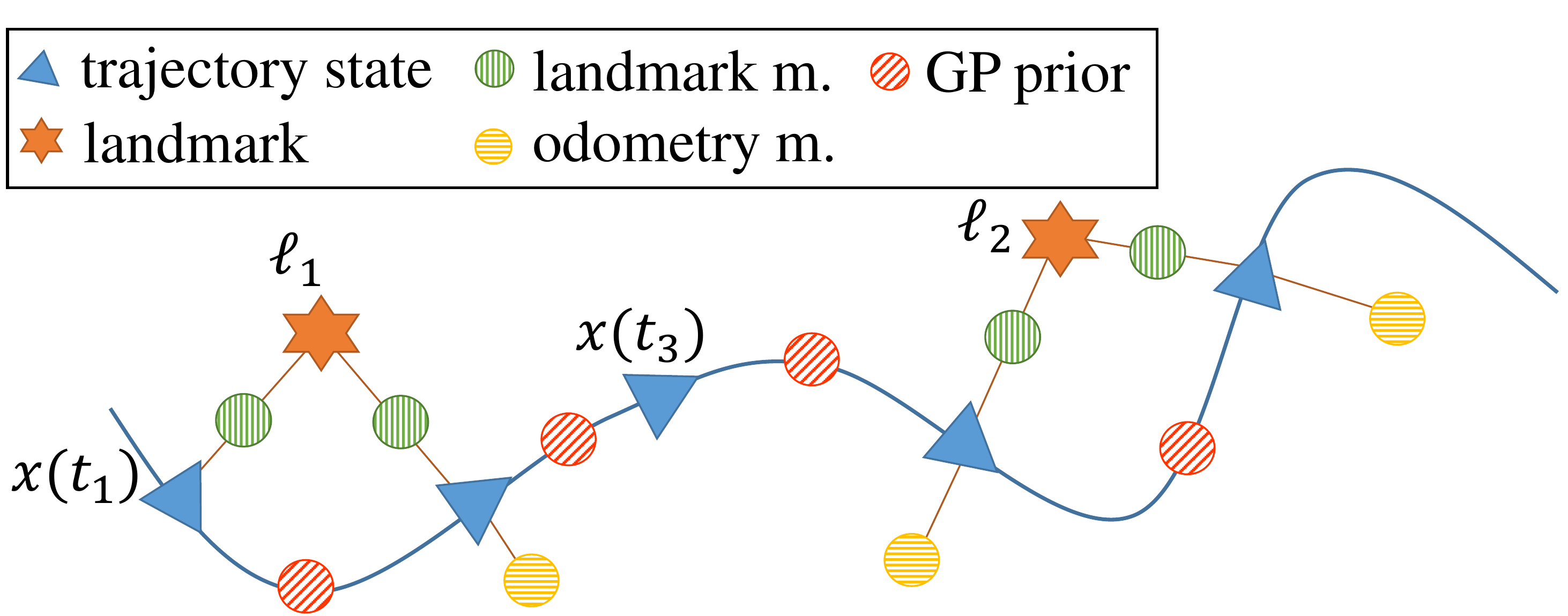}
\caption{A simple factor graph that includes landmark measurements, odometry measurements, and Gaussian process priors.} 
\label{fig_factor_graph}
\end{figure}

To understand how the Bayes tree is used, it is helpful to understand how the GP estimation problem can be represented as a factor graph~\cite{kschischang:2001}. Formally, a factor graph is a bipartite graph $G = (\mathcal{F}, \bm{\theta}, \mathcal{E})$, where $\mathcal{F}$ is the set of factor nodes that encodes all probabilistic constraints on variables, including landmark measurements, odometry measurements, and smoothing priors, $\bm{\theta}$ is the set of variable nodes to estimate, and $\mathcal{E}$ is the set of edges that connect factors nodes with variable nodes. 
The joint probability of variables to estimate is factored as
\begin{equation} \label{eqn_factorization}
f(\bm{\theta}) = \prod\limits_{i} f_i(\bm{\theta}_i)
\end{equation}
where $f_i \in \mathcal{F}$ is one of the factors, and $\bm{\theta}_i$ is the set of variables directly connected to $f_i$. The estimation problem is to find $\bm{\theta}^*$ that maximizes Eq.~\ref{eqn_factorization}. 
As stated in Section~\ref{subsec_sparse_GP_regression}, Barfoot
 et al. prove that when GPs are generated by LTV SDE, $\bm{\mathcal{K}}^{-1}$ is block-tridiagonal~\cite{Barfoot-RSS-14}. 
They also state that the factors resulted from the Gaussian process representation of the trajectory, or the Gaussian process prior factors, only connect consecutive pairs of states. This leads to a sparse Factor graph (Fig.~\ref{fig_factor_graph}). 

The factor graph can be converted to a Bayes net by a \emph{bipartite elimination game}~\cite{Heggernes96findinggood}. The procedure is equivalent to converting Eq.~\ref{eqn_map_linearized} to least squares form and computing the square-root information matrix $\bm{\mathcal{L}}$ via an incomplete Cholesky factorization. 
To facilitate marginalization and optimization, a \emph{Bayes tree}, which groups several variables together based on their dependence,  is constructed from the Bayes net~\cite{kaess2011bayes}. 
From a linear algebra perspective, the Bayes tree captures the structure of the Cholesky factor $\bm{\mathcal{L}}$ of $\bm{\mathcal{I}}$, and the sequence of back substitutions that can be performed.

When we add a new measurement, add a prior for a new variable, or relinearize a previous measurement, $\bm{\mathcal{L}}$ will change accordingly. However, all modifications to the factor graph only have \emph{local} effects on $\bm{\mathcal{L}}$. Exploiting this observation is the foundation for efficient incremental updates. 
Since the nodes of Bayes tree encode conditional probability distributions which directly correspond to rows in $\bm{\mathcal{L}}$, the structure of the tree can be leveraged to efficiently update the factor $\bm{\mathcal{L}}$~\cite{kaess2011bayes}. 

The nodes $\bm{\theta}_{nf}$ containing variables involved in new factors, or nodes $\bm{\theta}_{lin}$ whose linear step is larger than a predetermined threshold, are identified.
\footnote{The reader is referred to \cite{Kaess-iSAM2} for additional details regarding this just-in-time relinearization.}

 Only these nodes and their ascendants in the Bayes tree are then updated. 
When a sub-tree is updated, variables in the sub-tree are reordered by constrained COLAMD~\cite{Davis:2004} to ensure sparsity and heuristically maximize the locality of future updates. Finally, $\delta\bm{\theta}^*$ is computed from tree root to leaves. 
Propagation stops when updates to the conditioning variables are below a predetermined threshold.

Algorithm~\ref{alg_incremental_gpgn_bt} summarizes incremental Gaussian process regression with the Bayes tree data structure in detail. The algorithm also incorporates state interpolation, described in the next section.

\subsection{Faster Updates Through Interpolation} \label{sec_incremental_updates_interpolation}

To further reduce computation time, we take advantage of Gaussian process state interpolation (as suggested by~\citet{tong2013gaussian}) within our incremental algorithm. This allows us to reduce the total number of estimated states, while still using all of the measurements, including those that involve interpolated states. By only estimating a small fraction of the states along the trajectory, we realize a significant speedup relative to a naive application of the Bayes tree (see Section~\ref{sec_experiment}). This is an advantage of GP-based methods with respect to discrete-time methods like iSAM 2.0.

Algorithm~\ref{alg_incremental_gpgn_bt} describes how interpolation is used within our incremental algorithm: First, when a measurement related to a missing state is received, the variables necessary to interpolate the state, as well as the corresponding cliques in Bayes tree that should be removed or updated, are identified. Since the sparse GP has a  LTV SDE prior, each interpolated state is only a function of two nearby states (see Eq.~\ref{eqn_sparse_query}). These nearby states are therefore included into the set of variables $\bm{\theta}_{nf}$ related to the new factor (line 1). 
In the case that the GP relies on a different kernel matrix, the corresponding states used for interpolation can be determined from Eq.~\ref{eqn_batch_query}. Second, linearization of factors that involve missing states (line 3) is performed by incorporating state interpolation via Eq.~\ref{eqn_interpolated_measurement_linearization}.
\begin{algorithm}\footnotesize
\caption{Updating Sparse GP Regression by Bayes Tree} \label{alg_incremental_gpgn_bt}
\begin{algorithmic}
\WHILE{collecting data}
\vspace{2pt}
\STATE 1. Get new measurement results, store new factors $\mathcal{F}_{new}$ and identify related variables $\bm{\theta}_{nf} = \bigcup \bm{\theta}_{i}$, $f_i \in \mathcal{F}_{new}$. 
If the state $\bm{x}(t_i) \in \bm{\theta}_{nf}$ is missing, then it is replaced by variables used in interpolation (Eq.~\ref{eqn_sparse_query}); If $\bm{x}(t_i) \in \bm{\theta}_{nf}$ is a new state to estimate, the previous state to estimate is added to $\bm{\theta}_{nf}$, and a Gaussian process prior factor is stored.
\vspace{2pt}
\STATE 2. For each affected variable in $\bm{\theta}_{aff} = \bm{\theta}_{lin} \cup \bm{\theta}_{nf}$, remove the corresponding clique and ascendants up to the root of Bayes tree. 
\vspace{2pt}
\STATE 3. Relinearize the factors required to recreate the removed part. Use interpolation when linearizing factors involving missing states (Eq.~\ref{eqn_interpolated_measurement_linearization})
\vspace{2pt}
\STATE 4. Add cached marginal factors from orphaned sub-trees of removed cliques and create a factor graph
\vspace{2pt}
\STATE 5. Eliminate the factor graph by a new variable ordering, create a Bayes tree, and attach back orphaned sub-trees
\vspace{2pt}
\STATE 6. Partially update estimate from root to leaves and stop walking down a branch when the updates to variables that the child clique is conditioned on are not significant enough
\vspace{2pt}
\STATE 7. Collect variables involved in the measurement factors $\mathcal{F}_{lin}$ where previous linearization point is far from current estimate, $\bm{\theta}_{lin} = \bigcup \bm{\theta}_{i}$, $f_i \in \mathcal{F}_{lin}$
\vspace{2pt}
\ENDWHILE
\end{algorithmic}
\end{algorithm}

\section{Experimental Results} \label{sec_experiment}
We evaluate the performance of our incremental sparse GP regression algorithm to solving the STEAM problem on synthetic and real-data experiments and compare our approach to the state-of-the-art. 
In particular, we evaluate how variable reordering can dramatically speed up the batch solution to the sparse GP regression problem, and how, by utilizing the Bayes tree and interpolation for incremental updates, our algorithm can yield even greater gains in the online trajectory estimation scenario. We compare: 
\begin{itemize}
\item {\bf PB}: Periodic batch (described in Section \ref{sec_trajest}). This is the state-of-the-art algorithm presented in~\citet{Barfoot-RSS-14} (XL variable ordering), which is periodically executed as data is received.
\item {\bf PBVR}: Periodic batch with variable reordering (described in Section~\ref{sec_var_reordering}).
\item {\bf BTGP}: The proposed approach - Bayes tree with Gaussian process prior factors (described in Section \ref{sec_incremental_updates}).
\end{itemize}
If the GP is only used to estimate the state at measurement times, the proposed approach offers little beyond a reinterpretation of the standard discrete-time iSAM~2.0 algorithm. Therefore, we  also compare our GP-based algorithm, which leverages interpolation, to the standard Bayes tree approach used in iSAM~2.0.
We show that by interpolating large fractions of the trajectory during optimization, the GP allows us to realize significant performance gains over iSAM~2.0 with minimal loss in accuracy. For these experiments we compare: 
\begin{itemize}
\item{\bf without interpolation}: 
BTGP without interpolation at a series of lower temporal resolutions. Without interpolation BTGP is algorithmically identical to iSAM~2.0. Measurements between two estimated states are simply ignored. 
\item{\bf with interpolation}: 
BTGP with interpolation at a series of lower resolutions. In contrast to the above case, measurements between estimated states are fully utilized by interpolating missing states at measurement times (described in Section~\ref{sec_incremental_updates_interpolation}).
\item{\bf finest estimate}:
The baseline. BTGP at the finest resolution, estimating all states at measurement times. When measurements are synchronous with evenly-spaced waypoints and no interpolation is used, BTGP is identical to iSAM~2.0 applied to the full dataset.
\end{itemize}
All algorithms are implemented with the same C++ libray, GTSAM 3.2,\footnote{https://collab.cc.gatech.edu/borg/gtsam/} to make the comparison fair and meaningful. 
Evaluation is performed on three datasets summarized in Table~\ref{tab_experiments}. We first evaluate performance in a synthetic dataset (Section \ref{subsec_synthetic_experiment}), analyzing estimation errors with respect to ground truth data. Results using real-world datasets are then presented in Sections \ref{subsec_Autonomous_Lawnmower} and \ref{subsec_victoria_park_experiment}. 

\begin{table*}\small
\caption{Summary of experimental datasets}
\label{tab_experiments}
\begin{center}
\begin{tabular}{r ||r |r |r | r |r |r}
& \# time steps  & \# odo. m. & \# landmark m. & \# landmarks &  travel dist.(km) \\
\hline \hline
Synthetic  & 1,500 & 1,500 & 1,500 & 298 & 0.2 \\
\hline
Auto. Mower & 9,658 & 9,658 & 3,529 & 4 & 1.9 \\
\hline
Victoria Park & 6,969  & 6,969 &  3,640 &  151 & 3.5 \\
\end{tabular}
\end{center}
\end{table*}

\begin{figure}
\centering
  \includegraphics[width=.7\linewidth]{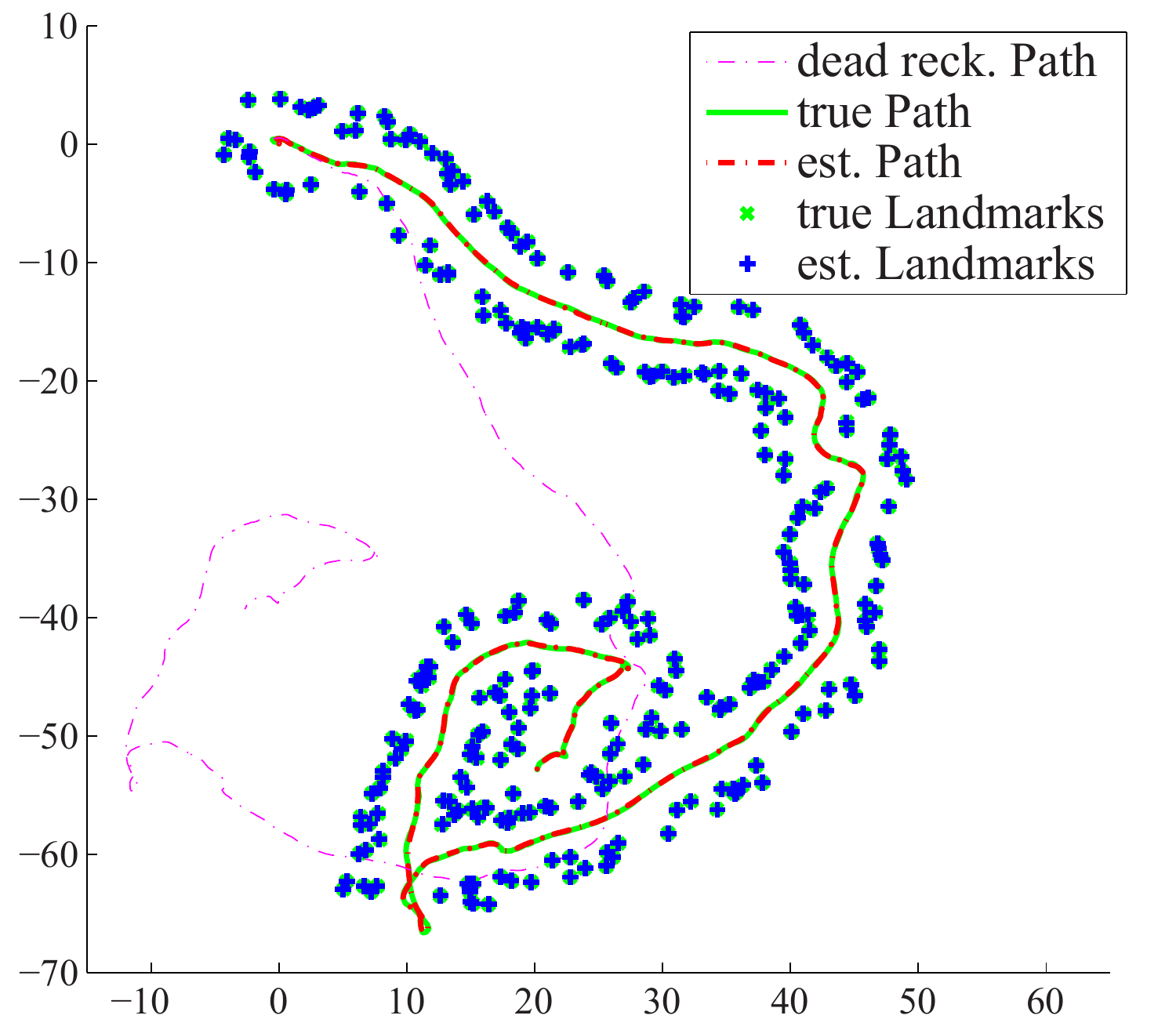}
\caption{Synthetic dataset: Ground truth and state estimates are shown. 
Lines between trajectory points and landmarks indicate range measurements. 
State estimates obtained from BTGP approach are very close to ground truth.}
\label{fig_syn_estimate}
\end{figure}

\subsection{Synthetic SLAM Exploration Task} \label{subsec_synthetic_experiment}
This dataset consists of an exploration task with 1,500 time steps. Each time step contains a trajectory state $\bm{x}(t_i) = [\begin{array}{cc} \bm{p}(t_i) & \dot{\bm{p}}(t_i) \end{array}]^\intercal$, $\bm{p}(t_i) = [\begin{array}{ccc} x(t_i) & y(t_i) & \theta(t_i) \end{array}]^\intercal$, an odometry measurement, and a range measurement related to a nearby landmark. The total number of landmarks is 298. The trajectory is randomly sampled from a Gaussian process generated from white noise acceleration $\ddot{\bm{p}}(t) = \bm{w}(t)$, i.e. constant velocity, and with zero mean.
\begin{equation} \label{eqn_syn_process}
\dot{\bm{x}}(t) = \bm{A} \bm{x}(t) + \bm{F} \bm{w}(t)
\end{equation}
where
\begin{align}
\bm{x}(t) &= 
\begin{bmatrix}
  \bm{p}(t) \\
  \dot{\bm{p}}(t)
\end{bmatrix}
, \hspace{12pt}
\bm{p}(t) = 
\begin{bmatrix}
  x(t) \\
  y(t) \\
  \theta(t)
\end{bmatrix}
, \hspace{12pt}
\bm{A} = 
\begin{bmatrix}
 \bm{0}& \bm{I} \\
 \bm{0}& \bm{0} 
\end{bmatrix}
, \nonumber\\
\bm{F} &= 
\begin{bmatrix}
 \bm{0}\\
 \bm{I}
\end{bmatrix}
, \hspace{12pt}
\bm{w}(t) \sim \mathcal{GP} (\bm{0}, \bm{Q}_c\delta(t-t'))
\end{align}
Note that velocity $\dot{\bm{p}}(t)$ must to be included in trajectory state to represent the motion in LTV SDE form~\cite{Barfoot-RSS-14}. 

The odometry and range measurements with Gaussian noise are specified in Eq.~\ref{eqn_odometry_measurements} and Eq.~\ref{eqn_range_measurements} respectively.
\begin{equation} \label{eqn_odometry_measurements}
\bm{y}_{io} =
\begin{bmatrix}
\cos\theta(t_{i})\cdot \dot{x}(t_{i}) + \sin\theta(t_{i})\cdot \dot{y}(t_{i}) \\
\dot{\theta}(t_{i})
\end{bmatrix}
+ \bm{n}_o
\end{equation}
\begin{equation} \label{eqn_range_measurements}
y_{ir} = \left\|
\begin{bmatrix}
x(t_i) &
y(t_i)
\end{bmatrix}^{\intercal}
 - \bm{\ell}_{j} \right\|_2 + n_r
\end{equation}
where $\bm{y}_{io}$ consists of the robot-oriented velocity and heading angle velocity with Gaussian noise, and $y_{ir}$ is the distance between the robot and a specific landmark $\bm{\ell}_{j}$ at $t_i$ with Gaussian noise. 

\begin{figure*}
\centering
  \includegraphics[width=.95\linewidth]{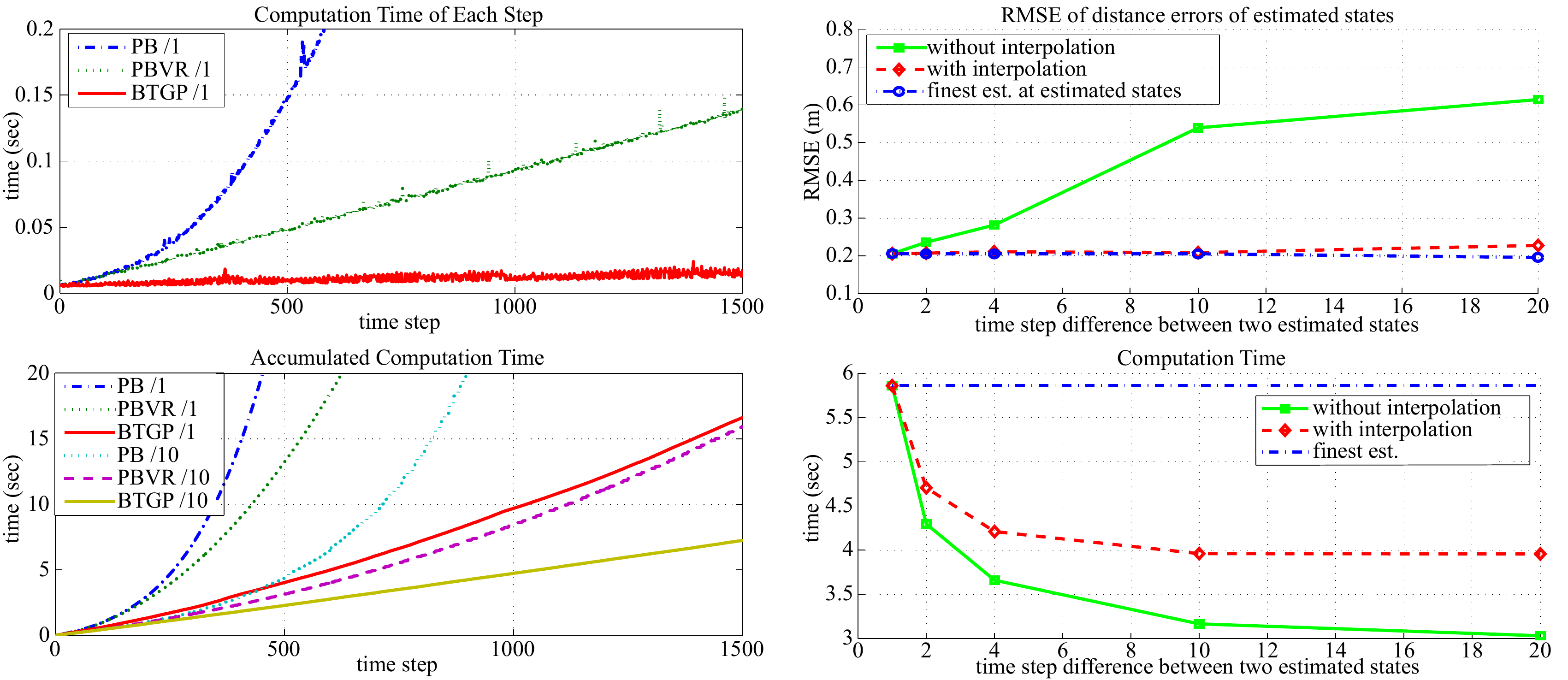}
  \caption{Synthetic dataset: {\bf (Left Column)} Comparison of the computation time of three approaches PB, PBVR, and BTGP. The modifiers /1 and /10 indicate frequency of state updates. For example: \emph{BTGP/1} updates the estimate after 1 new range measurement using BTGP. Likewise  \emph{BTGP/10} updates the estimate after 10 new range measurements using BTGP. For fair comparison no interpolation is used by BTGP. Due to the large number of landmarks,  298, compared to the number of trajectory states, variable reordering dramatically improves the performance. {\bf (Right Column)} Trade-off between computation time and accuracy if BTGP makes use of interpolation. The $y$-axis measures the RMSE of distance errors of the estimated trajectory states and total computation time with increasing amounts of interpolation. The $x$-axis measures the time step difference between two estimated (non-interpolated) states. ``Without interpolation'' means that the number of states is reduced, but measurements taken at the missing states are ignored. This is equivalent to running iSAM~2.0 to find a trajectory with coarse discretization. ``With interpolation'' is the BTGP algorithm that interpolates missing states while incorporating odometry measurements at the interpolated states. ``Finest estimate'' is the baseline which measures RMSE and computation time if the number of states is not reduced. This is exactly equivalent to iSAM~2.0 run on the full measurement and odometry data. The results indicate that interpolating $\sim90\%$ of the states (i.e. estimating only $\sim10\%$ of the states) while running BTGP can result in a $33\%$ reduction in computation time over iSAM~2.0 without sacrificing accuracy. } 
\label{fig_syn_performance}
\end{figure*}

We compare the computation time of the three approaches (PB, PBVR and BTGP) in Fig.~\ref{fig_syn_performance}. The incremental Gaussian process regression (BTGP) offers significant improvements in computation time compared to the batch approaches (PBVR and PB). 

We also demonstrate that BTGP can further increase speed over a naive application of the Bayes tree (e.g. iSAM~2.0) without sacrificing much accuracy by leveraging interpolation. 
To illustrate the trade-off between the accuracy and time efficiency due to interpolation,  we plot RMSE of distance errors and the total computation time by varying the time step difference (the rate of interpolation) between estimated states.

\begin{figure}
\centering
  \includegraphics[width=0.7\linewidth]{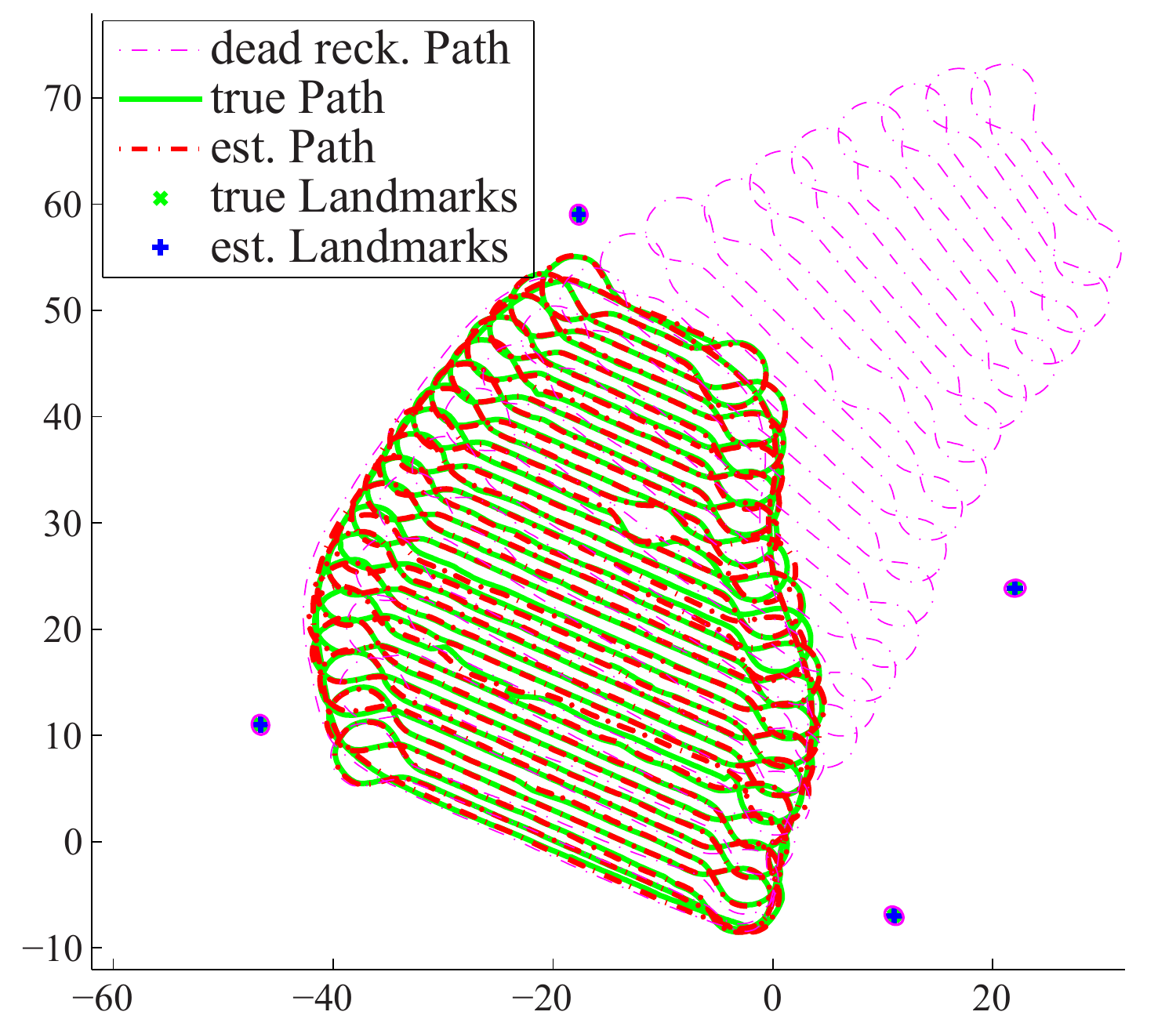}
\caption{The Autonomous Lawnmower dataset: Ground truth and state estimates are shown. The range measurements are sparse, noisy, and  asynchronous. Ground truth and state estimates obtained from BTGP are very close.}
\label{fig_plaza_estimate}
\end{figure}

\subsection{The Autonomous Lawnmower}\label{subsec_Autonomous_Lawnmower}
The second experiment evaluates our approach on real data from a  freely available range-only SLAM dataset collected from an autonomous lawn-mowing robot~\cite{Djugash2010}.  The ``Plaza'' dataset consists of odometer data and range data to stationary landmarks collected via time-of-flight radio nodes. (Additional details on the experimental setup can be found in~\cite{Djugash2010}.)  Ground truth paths are computed from GPS readings and have 2cm accuracy according to~\cite{Djugash2010}.  

The environment, including the locations of the landmarks and the ground truth paths, are shown in
Fig.~\ref{fig_plaza_estimate}.  
The robot travelled 1.9km, occupied 9,658 poses, and received 3,529 range measurements, while following a typical path generated during mowing. The dataset has sparse range measurements, but contains odometry measurements at each time step. 
The results of incremental BTGP are shown in Fig.~\ref{fig_plaza_estimate} and demonstrate that we are able to estimate the robot's trajectory and map with a very high degree of accuracy.

\begin{figure*}
\centering
  \includegraphics[width=.9\linewidth]{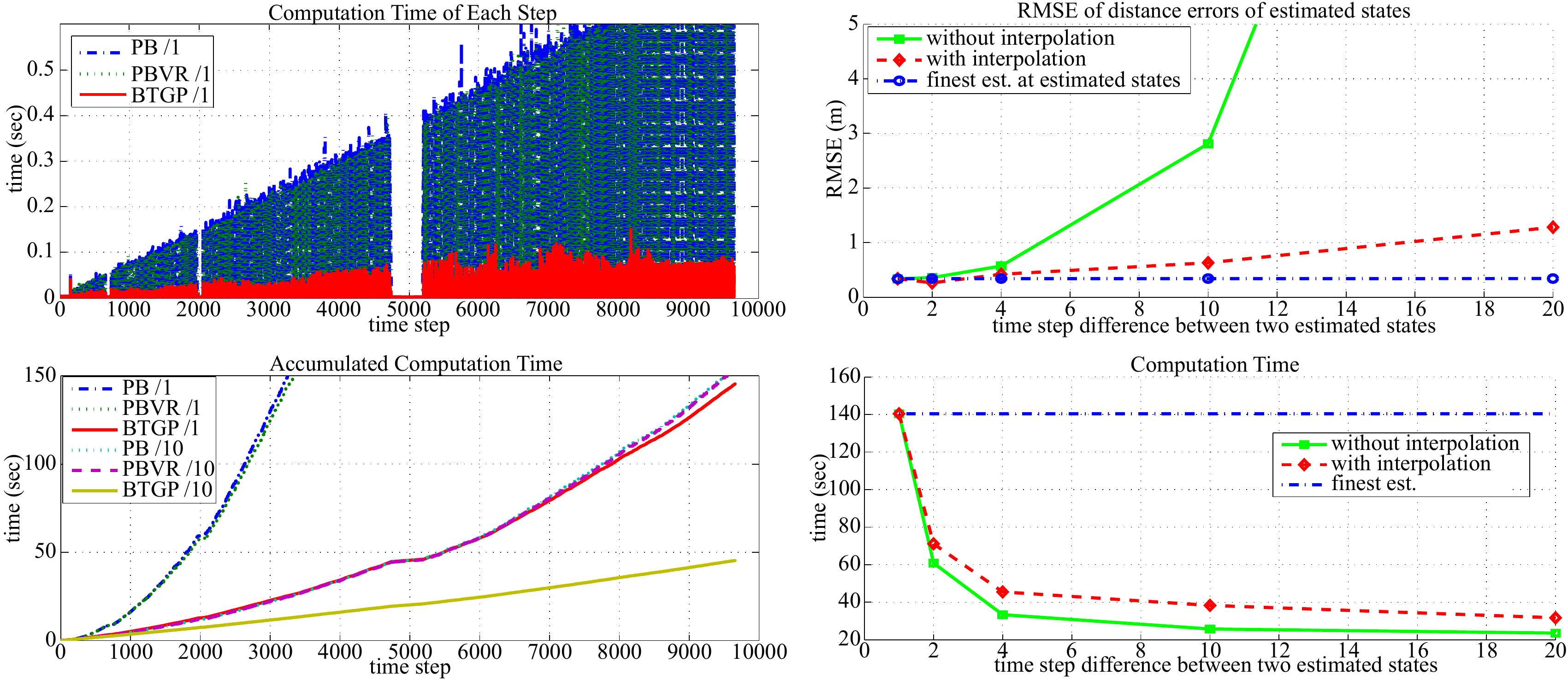}
  \caption{Autonomous Lawnmower dataset: ({\bf Left Column}) Comparison of the computation time of three approaches PB, PBVR, and BTGP. As in Figure~\ref{fig_syn_performance}, the modifiers /1 and /10 indicate frequency of state updates. For fair comparison no interpolation is used by BTGP in these experiments. Updates are very fast (close to zero time) when there are no range measurements. This dataset has no range measurements for a long, sustained stretch around the 5000th tilmestep, which accounts for the `gap' in the upper left-hand graph. Due to the low number of landmarks, variable reordering does not help (PB and PBVR take roughly the same amount of time). The incremental BTGP approach dramatically reduces computation time. ({\bf Right Column}) Trade-off between computation time and accuracy if BTGP makes use of interpolation. The $y$-axis measures the RMSE of distance errors of the estimated trajectory states and total computation time with increasing amounts of interpolation. The $x$-axis measures the time step difference between two estimated (non-interpolated) states. ``Without interpolation'' means that the number of states are reduced, but measurements taken at the missing states are ignored. ``With interpolation'' is the standard BTGP algorithm that interpolates missing states. ``Finest estimate'' is the baseline which measures RMSE and computation time if no states are interpolated, which is exactly equivalent to running iSAM~2.0 on the full set of measurements and odometry. The results indicate that interpolating $\sim80\%$ of the states within BTGP results in only an 8cm increase in RSME while reducing the overall computation time by $68\%$ over iSAM~2.0.} 
\label{fig_plaza_performance}
\end{figure*}

As in Section~\ref{subsec_synthetic_experiment}, performance of three approaches -- periodic batch relinearization (PB), periodic batch relinearization with variable reordering (PBVR) and incremental Bayes tree (BTGP) are compared in Fig.~\ref{fig_plaza_performance}. 

In this dataset, the number of landmarks is 4, which is extremely small relative to the number of trajectory states, so there is no performance gain from reordering. However, the Bayes tree-based approach dramatically outperforms the other two approaches. As the problem size increases, there is negligible increase in computation time, even for close to 10,000 trajectory states.

\subsection{Victoria Park} \label{subsec_victoria_park_experiment}
The third experiment evaluates our approach on the Victoria Park dataset~\cite{Guivant:2001}, which consists of range-bearing measurements to landmarks, and speed and steering odometry measurements. The data was collected from a vehicle equipped with a laser sensor driving through the Sydney's Victoria Park. The environment contains a high number of trees as landmarks. The vehicle travelled $\sim3.5$ km in 26 minutes. After repeated measurements, taken when the vehicle is stationary, are dropped, the dataset consists of 6,969 time steps and 3,640 range-bearing measurements relative to 151 landmarks. The bearing measurement is specified in Eq.~\ref{eqn_bearing_measurements}, as the relative angle from vehicle heading to the landmark direction with Gaussian noise  where $[\begin{array}{cc} x_{j} & y_{j} \end{array}]^\intercal$ is location of landmark $j$.
\begin{equation} \label{eqn_bearing_measurements}
y_{ib} = \text{atan2} \left(y_{j} -  y(t_i), x_{j} - x(t_i)\right) - \theta(t_i) + n_{ib}
\end{equation}

\begin{figure*} 
\centering
 \includegraphics[width=.9\linewidth]{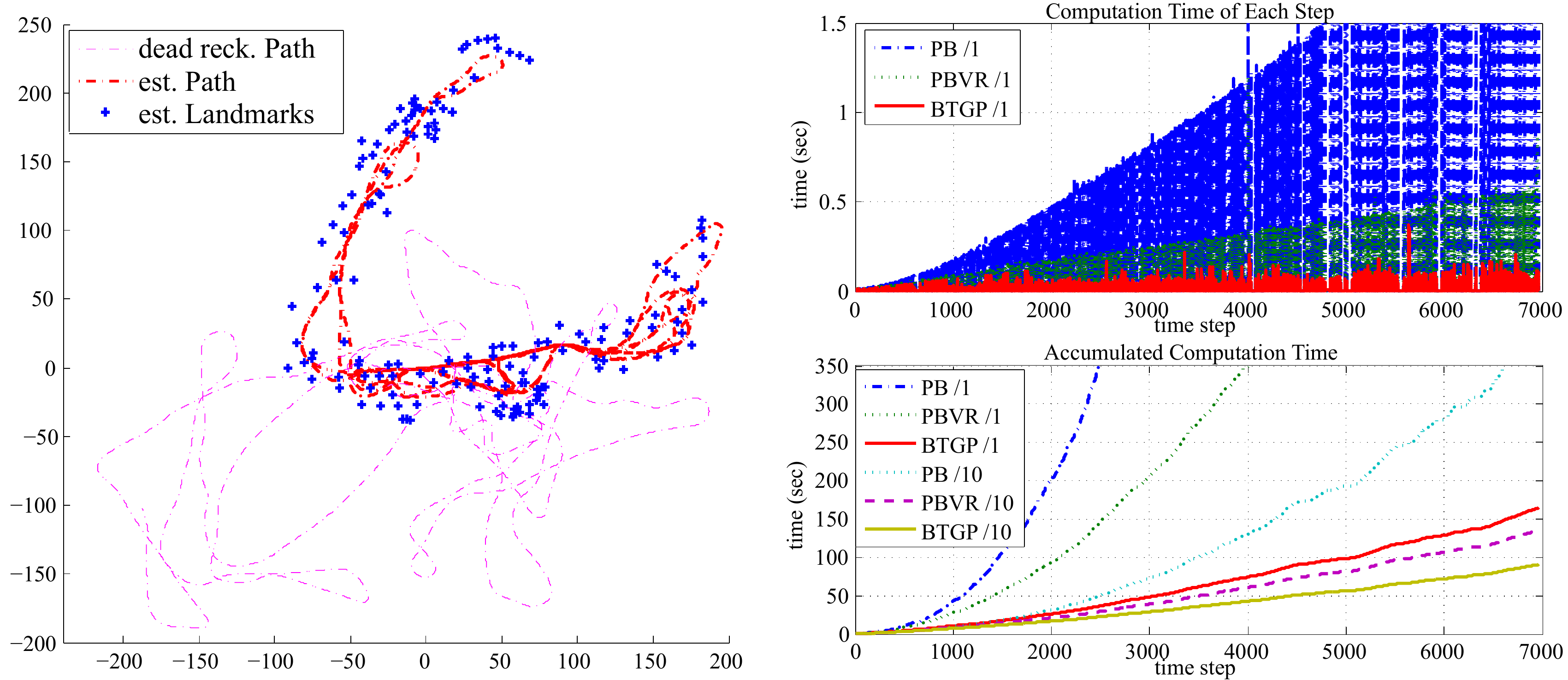}
\caption{Victoria Park dataset: (left) dead reckoning and estimated path obtained from BTGP approach. (right)  Comparison of the computation time of three approaches PB, PBVR, and BTGP. As in Figures~\ref{fig_syn_performance} and~\ref{fig_plaza_performance}, the modifiers /1 and /10 indicate frequency of state updates. Updates are very fast (close to zero time) when there are no range measurements. Since many landmarks are involved, PBVR dramatically improves performance, compared PB. The incremental BTGP algorithm improves performance even further. Unlike in previous datasets, we did not evaluate the trade-off between interpolation and accuracy for Victoria Park, since we do not have access to ground truth and cannot evaluate the effect on accuracy. However, like previous datasets, interpolation can greatly increase the speed of BTGP.}
\label{fig_victoria_park}
\end{figure*}

The results, shown in Figure \ref{fig_victoria_park}, further demonstrate the advantages of BTGP. As seen from the upper right plot, variable reordering drastically reduces computation time when used within batch optimization (PBVR), and even further in the incremental algorithm (BTGP).

\section{Conclusion}
We have introduced an incremental sparse Gaussian process regression algorithm for computing the solution to the continuous-time simultaneous trajectory estimation and mapping (STEAM) problem. The proposed algorithm elegantly combines the benefits of Gaussian process-based approaches to STEAM while simultaneously employing state-of-the-art innovations from incremental discrete-time algorithms for smoothing and mapping. 
Our empirical results show that by parameterizing trajectories with a small number of  states and utilizing Gaussian process interpolation, our algorithm can realize large gains in speed over iSAM~2.0 with very little loss in accuracy (e.g. reducing computation time by $68\%$ while increasing RMSE by only 8cm on the Autonomous Lawnmower Dataset) .

\bibliographystyle{plainnat}

\end{document}